# Machine Learning Algorithms for Prediction of Penetration Depth and Geometrical Analysis of Weld in Friction Stir Spot Welding Process

**Raheem Al-Sabur[1], Akshansh Mishra[2] and Ahmad K. Jassim[3]**

[1] Department of Mechanical Engineering, University of Basrah, Basra 61001, Iraq
[2] Materials Engineering and Nanotechnolgy, Politecnico Di Milano, Milan, Italy
[3] Department of Materials Engineering, University of Basrah, Basra 61001, Iraq

**ABSTRACT:**

Nowadays, manufacturing sectors harness the power of machine learning and data science algorithms to make predictions for the optimization of mechanical and microstructure properties of fabricated mechanical components. The application of these algorithms reduces the experimental cost beside leads to reduce the time of experiments. The present research work is based on the prediction of penetration depth using Supervised Machine Learning algorithms such as Support Vector Machines (SVM), Random Forest Algorithm, and Robust Regression algorithm. A Friction Stir Spot Welding (FSSW) was used to join two elements of AA1230 aluminum alloys. The dataset consists of three input parameters: Rotational Speed (rpm), Dwelling Time (seconds), and Axial Load (KN), on which the machine learning models were trained and tested. It observed that the Robust Regression machine learning algorithm outperformed the rest of the algorithms by resulting in the coefficient of determination of 0.96. The research work also highlights the application of image processing techniques to find the geometrical features of the weld formation.

## 1. INTRODUCTION

Machine Learning is a sub-branch of Artificial Intelligence that uses mathematical or statistical algorithms to extract the required available pattern from the data. For extracted patterns from the available data, the study of Machine Learning generally deals with the uncovering of information hiding behind these patterns [1-3]. It should be noted that Machine Learning, Deep Learning, Artificial Intelligence, and Data Science are interrelated to each other as shown in figure 1. It is observed that the human-intensive tasks can be replaced by the automated solutions provided by Artificial Intelligence algorithms.

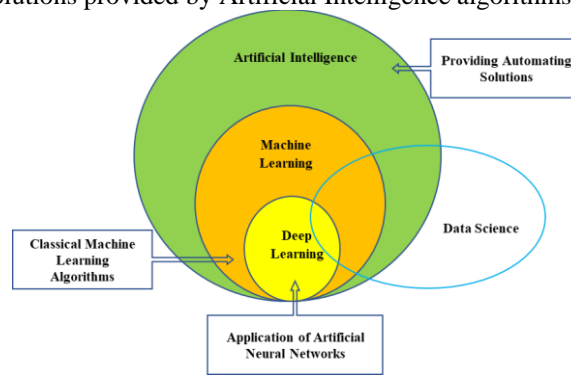

Figure 1. The relationship between, Machine Learning, Deep Learning, Artificial Intelligence, and Data Science

Based on the style and method involved, Machine Learning algorithms are divided into four types:, Reinforcement Learning, Unsupervised Learning, Supervised Machine Learning, and Semi-Supervised Learning [4, 5]. Supervised Machine Learning algorithm deals with the prediction of the unseen labeled data based on known labeled data. Unsupervised Learning is based on the self-discovery of hidden patterns from the unlabeled data. Reinforcement Learning-based algorithms work on trial-and-error analysis simulated in an interactive virtual environment.

Friction stir spot welding (FSSW) process is a creative solid-state welding method invented in the 1990s for joining two sheets. FSSW process is depending on the heat generated from the friction contact to produce the required plastic deformation. It is environmentally friendly technology, a non-consumable tool, and has no type of slag, no shielding gas, and no fumes[6, 7]. In FSSW, the optimization of mechanical properties is widely used to find the best values of welding parameters such as rotational speed, traverse speed, axial load , tool geometry ...etc. [8].

Machine Learning algorithms have been used for various applications in the welding process. This algorithm was applied as an approach for predicting the thermal field of the multi-paths in gas metal arc welding [9]. Support Vector Machine (SVM) and Decision Trees techniques were used for classifying the weld defects in the gas metal arc welding [10]. Artificial Neural Network (ANN) algorithm was used in small-scale resistance spot welding analyses for predicting the contact resistance of zirconium joints [11]. The Supervised Learning regression-based algorithms and Supervised Learning classification-based algorithms have been used for predicting the efficiency of the friction stir welding joint efficiency and the ultimate tensile strength [12]. The welding efficiency in similar friction stir welded copper joints were detected using Machine Learning-based classification algorithms. It observed that the Artificial Neural Network model successfully predicted the welding efficiency with an accuracy score of 94 percent [13]. The Machine Learning-based image processing approach was used for determining the microstructure grains size in the friction stir welded AZ31B alloy plate [14]. Artificial Neural Network (ANN) algorithm predicted the corrosion potential in Friction Stir Welded joints with good accuracy [15]. Balachandar et al. [16] used Random Forest-based Machine Learning algorithm for condition monitoring of the Friction Stir Welding Tool. Du et al. [17] used machine learning algorithms to analyze 114 sets of experimental data of the tool failure to estimate the hierarchy of causative variables. In general, related studies on the application of Machine Learning algorithms in friction stir spot welding are limited.

In the present study, three types of Supervised Machine Learning-based regression algorithms have been used: Robust Regression, Random Forest, and Support Vector Machines (SVM) for predicting the maximum penetration depth, and for geometrical analysis of the weld shape, image processing tools have been used which will be discussed in upcoming sections.

## 2. METHODOLOGY
### 2.1. Support Vector Regression Algorithm

The main objective of the simple linear regression models is the minimization of the sum of squared errors. Equation 1 shows the expression for an objective function of the ordinary least squares (OLS) with one feature or predictor [18].

$$MIN \sum_{i=1}^{n}(y_i - w_i x_i)^2 \qquad (1)$$

Where, $y_i$ is the target variable, $w_i$ is the coefficient and $x_i$ is the feature.

Support Vector Regression algorithm provides the flexibility to provide a proper definition on the acceptance of a range of errors in the implemented Machine Learning model and further gives a hyperplane in hyper dimensions which is an appropriate line to fit the data. The objection function of the Support Vector Regression algorithm is minimizing the coefficients of the l2-norm of the coefficient vector as shown in Equation 2. The constraints handle error terms as shown in Equation 3, where the absolute error is set less than or equal to the specified margin, which is called maximum error. To gain the desired accuracy of the model, the maximum error can be tuned as below [18].

$$MIN \frac{1}{2}\|w\|^2 \qquad (2)$$
$$|y_i - w_i x_i| \leq \varepsilon \qquad (3)$$

### 2.2. Random Forest Algorithm

Random Forest is a Supervised Machine Learning algorithm that uses the collection learning method for making predictions of the output variables, i.e., to result in more accurate output prediction than a single model, it sums up the predictions yielded by various machine learning algorithms. The structure of Random Forest is shown in figure 2.

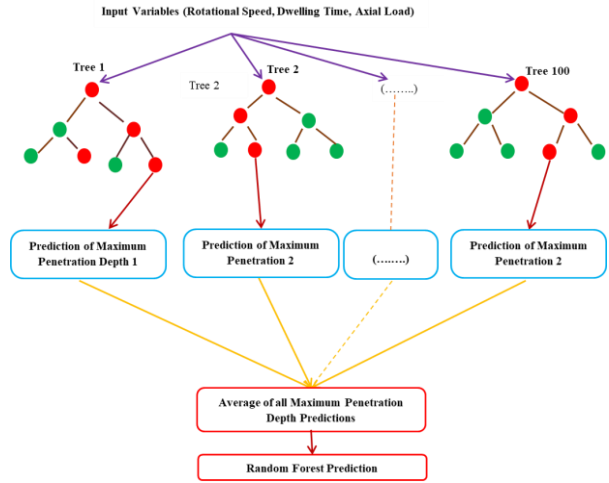

Figure 2: Structure of Random Forest Algorithm

### 2.3. Robust Regression Algorithm

It should be kept in mind that the estimation of ordinary least squares is optimal for the linear regression algorithm when all the regression assumptions considered are valid. Generally, the least square regression performs poorly if some of these assumptions are invalid. A Robust Regression algorithm requires less restrictive assumptions and, thus it is the best alternative to least squares regression. For the implementation of Robust Regression, we have considered the size of the dataset, i.e., n that is equal to 27 such that:

$$y_i = x_i^T \beta + \epsilon_i \quad (4)$$
$$\epsilon_i(\beta) = y_i - x_i^T \beta \quad (5)$$

Where i=1, 2, 3, ……….27 and in Equation 5 describes the error term's dependency on the regression coefficients. It is observed that the obtained results are the best line of fit by using the Robust Regression algorithm which is yielding the $R^2$ value of 0.963 on the testing set.

### 2.4. Image Processing Algorithm

Any image can be considered as a function of two variables which are spatial i.e., where the Cartesian location represents the brightness. A finite number of elements known as pixels located at a particular point constitute the digital image as shown in figure 3.

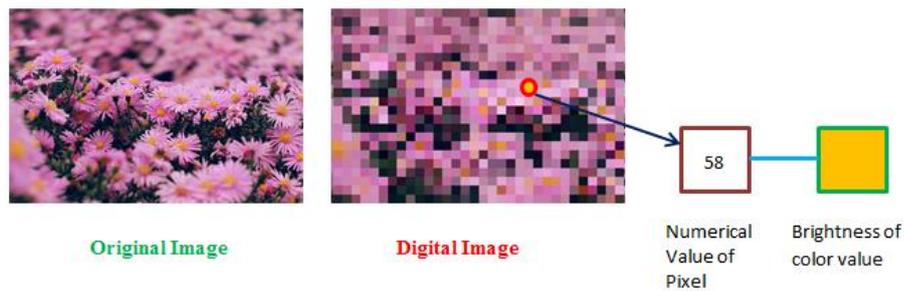

Figure 3: Representation of the digital image in terms of pixels

### 3. EXPERIMENTAL PROCEDURE

Aluminum alloy AA1230 specimens (3 x 25 x 100 mm) were lap joint welded by FSSW in this study. The chemical composition of the AA1230 alloy is shown in table 1.

Table 1. Chemical compositions of AA1230 Al-alloy

| element | Si | Fe | Mn | Mg | Cu | Al |
|---|---|---|---|---|---|---|
| % | 0.110 | 0.570 | 0.01 | 0.001 | 0.020 | balance |

A tool of tungsten carbide material was used, 55mm shoulder-length, 10mm shoulder diameter,

5mm pin length, 3mm pin diameter with a tilt angle of 12º. A load cell was used to measure the axial force with real-time monitoring using Labjack3 ADC data transmitter as shown in the figure 4.

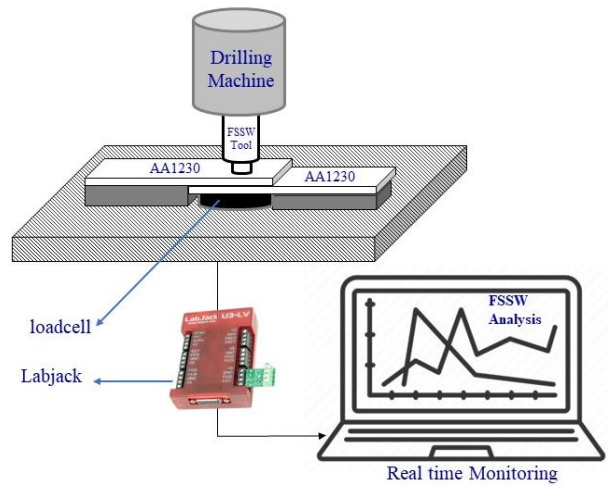

Figure 4: Real time monitoring and the specimen setup

In the present work, the main objective is to predict the maximum penetration depth (mm) based on its attributes which are rotational speed (rpm), dwelling time (sec), and axial force (kN). Table 2 shows the experimental dataset used in the current research work joined by FSSW process.

Table 2. Structure of dataset to predict the Maximum Penetration Depth

| Rotational Speed (rpm) | Dwelling Time (s) | Axial Force (kN) | Maximum Penetration (mm) |
|---|---|---|---|
| 1000 | 10 | 110 | 3.5 |
| 1000 | 10 | 120 | 4.01 |
| 1000 | 10 | 130 | 4.65 |
| 1000 | 20 | 110 | 3.96 |
| 1000 | 20 | 120 | 4.47 |
| 1000 | 20 | 130 | 4.92 |
| 1000 | 30 | 110 | 4.43 |
| 1000 | 30 | 120 | 5.02 |
| 1000 | 30 | 130 | 5.28 |
| 1500 | 10 | 110 | 3.88 |
| 1500 | 10 | 120 | 4.45 |
| 1500 | 10 | 130 | 5.15 |
| 1500 | 20 | 110 | 4.51 |
| 1500 | 20 | 120 | 5.02 |
| 1500 | 20 | 130 | 5.51 |
| 1500 | 30 | 110 | 4.64 |
| 1500 | 30 | 120 | 5.17 |
| 1500 | 30 | 130 | 5.66 |
| 2000 | 10 | 110 | 4.29 |
| 2000 | 10 | 120 | 4.79 |
| 2000 | 10 | 130 | 5.33 |
| 2000 | 20 | 110 | 4.88 |
| 2000 | 20 | 120 | 5.39 |
| 2000 | 20 | 130 | 5.87 |
| 2000 | 30 | 110 | 5.15 |
| 2000 | 30 | 120 | 5.71 |
| 2000 | 30 | 130 | 6.00 |

For understanding the metal flow in FSSW and classifying the holes shape, a high-resolution camera was used to capture the welding holes. The images were divided into three groups according to the tool rotational speed: 1000, 1500, and 2000 rpm as shown in figure 5-7.

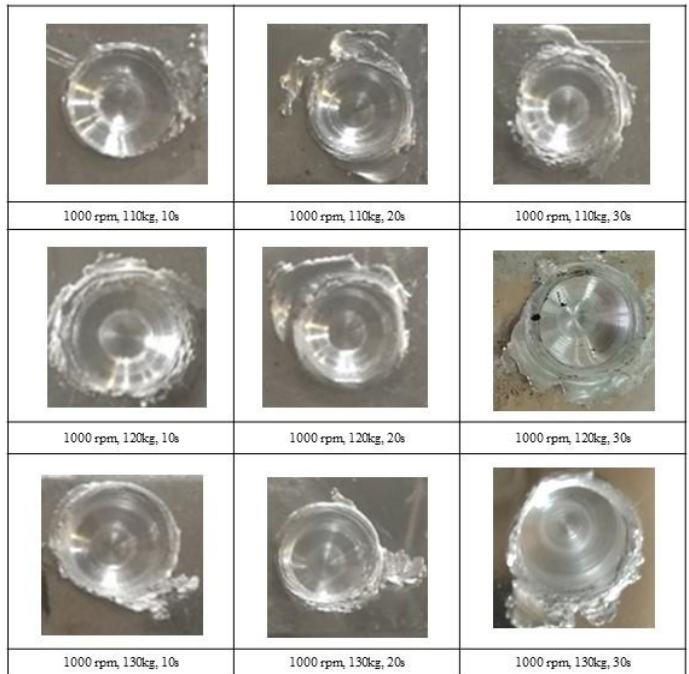

Figure 5: Holes shape at the rotation speed of 1000 rpm

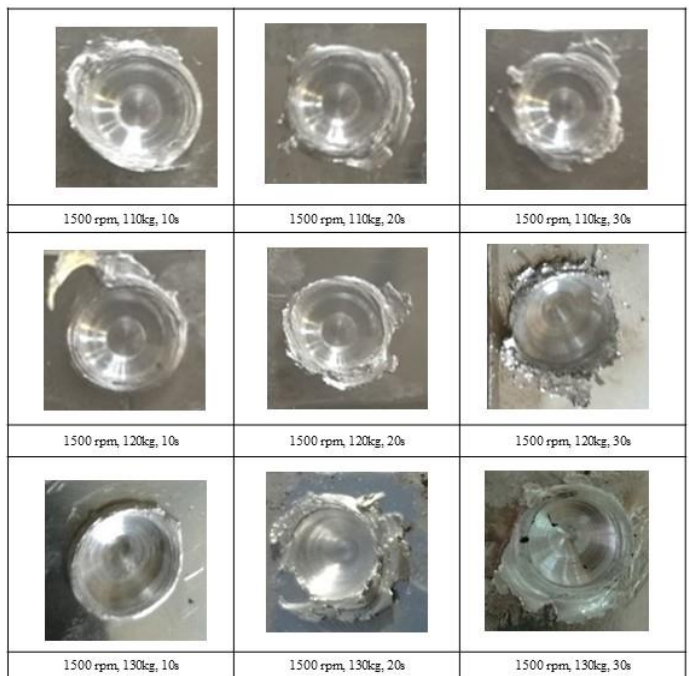

Figure 6: Holes shape at the rotation speed of rpm

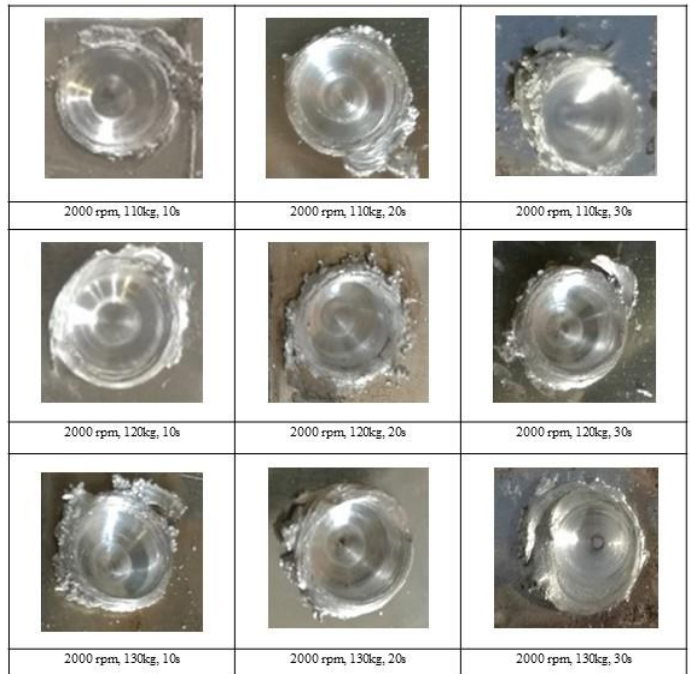

Figure 7: Holes shape at the rotation speed of 2000 rpm

## 4. FINDINGS AND DISCUSSION

In order to extract the hidden pattern, Machine Learning based statistical algorithms are shown historical data. So, this process of learning from the available historical data is called a 'training' Machine Learning algorithm. It should be noted that the training data contains both input and output variables. The Machine Learning algorithm establishes a mathematical relationship between the output variable and input variables during the training phase. So, the main objective of generating the relationship between output and input variables is to obtain a required mathematical equation that can predict the output variable by using unknown input variables. In the dataset, the output variable is named the dependent variable or target variable, and input variables are called independent variables.

The vector space diagram for the same dataset will look like figure 8. In this case, each row or training attribute is a feature vector or an array.

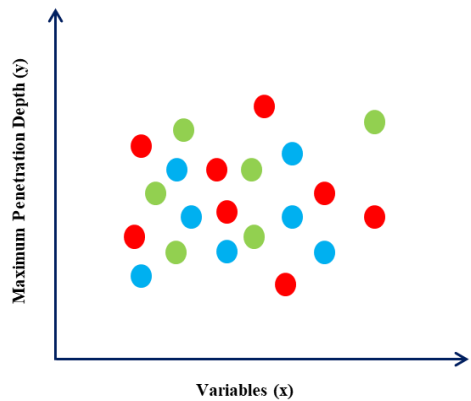

Figure 8: Representation of Maximum Penetration Depth and other variables in a vector space diagram

The next step is to begin the training process and the main objective is to attain a mathematical function for its optimization by carrying out various iterations i.e., to increase the accuracy in predicting the maximum penetration depth. Mathematically, the function of maximum penetration depth can be represented by Equation 6.

$$maximum\ penetration\ depth = f(rotaional\ speed, dwelling\ time, axial\ load) \qquad (6)$$

So, the main objective of the present work implementing the Supervised Machine Learning algorithms is to achieve Equation 6. In the experimental dataset and in the given equation, the maximum penetration depth is the target variable while rotational speed, dwelling time, and axial load are independent variables. In figure 9, the maroon line indicates the mathematical function or the Machine Learning equation which is also called the line of the best fit. The objective goal of this work is to arrive nearest to this mathematical function.

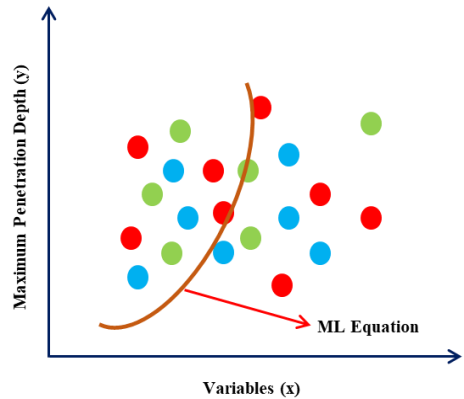

Figure 9: Machine Learning Equation using regression in a vector-space diagram.

The statistical modelling is performed by splitting the experimental dataset into 80:20 ratio as train and test data. The second part of the research work deals with the estimation of geometrical features of the weld shapes at different input parameters shown in figures 5-7.

### 4.1. Prediction of Maximum Penetration Depth

The performance of the implemented Supervised Machine Learning Algorithms is evaluated by the calculation of metric features such as Mean Square Error (MSE), Mean Absolute Error (MAE), and Coefficient of Determination, i.e., R-squared value. Mean Absolute Error measures the average absolute difference between actual and predicted values. Mean Square Error measures the average of the squared difference between the actual and predicted values in the given dataset. R squared value measures the closeness of the data which are fitted to the regression line. Table 3 shows the obtained results by implementing Supervised Machine Learning algorithms. It is observed that the Robust Regression algorithm yields a good fitting of the curve as it resulted in the highest value of 0.963 for the testing dataset while Support Vector Machine Algorithm is the second-best algorithm which resulted from the value of 0.895 on the testing dataset.

Table 3. Evaluation of testing and training sets using different Machine Learning algorithms

| Method | Testing sets | | | | Training sets | | | |
|---|---|---|---|---|---|---|---|---|
| | MAE | MSE | RMSE | $R^2$ | MAE | MSE | RMSE | $R^2$ |
| Support Vector Machine | 0.160 | 0.048 | 0.219 | 0.895 | 0.001 | 1.193 | 0.001 | 0.999 |
| Random Forest | 0.241 | 0.104 | 0.323 | 0.771 | 0.090 | 0.011 | 0.106 | 0.967 |
| Robust Regression | 0.124 | 0.016 | 0.128 | 0.963 | 0.081 | 0.008 | 0.092 | 0.975 |

### 4.2. Geometrical Features Analysis Using Image Processing

Welding geometrical features analysis is done by loading the given set of images in the Google Collaboratory platform and the extraction of statistical features is done with the help of Python programming. First, the given RGB images of weld shapes are converted to grayscale images. To track the size of a pixel while working on the weld images, a particular scale is defined. Second, the denoising process is carried out to obtain the threshold image to separate the boundaries of the weld from its surroundings. Third, image clean-up is carried out and a mask is created for visualizing the zones covered by welding shapes. The fourth step is to label the regions obtained from the masked image. The last step is to measure the geometrical features of the weld and import available data in the format of an Excel spreadsheet.

The Python libraries which were imported for subjecting the images of the welding samples to image processing algorithm are NumPy, cv2, pyplot from matplotlib, io, colour for loading the images to the color library. Next, assign the different color patches to a particular region, and then, a standard scale is defined where 1 pixel is equal to 1 micrometer. First, to initiate the thresholding, it must be observed the histograms of the weld sample 6 as shown in figure 10 a). The images are a two-dimensional array, while, on the other hand, the histograms are a one-dimensional array. So, there is a need to flatten the image i.e., convert a two-dimensional array into a one-dimensional array. The one-dimensional array bins are equal to 100 and the range is lying between (0-255).

From figure 10 a), a bunch of pixels is observed as lying between 100 to 200. The thresholding operation is further divided into two types, i.e., manual thresholding and auto thresholding. The thresholded value obtained for the sixth weld sample is equal to 160. The obtained thresholded image for the first weld sample is shown in figure 10 b).

It is observed that all pixels with the region of interest in the weld zone area of values is 255 while outside the weld region pixel values will be 0. It should be noted that the obtained image is thresholded, not binary. So, the conversion of thresholded image to the binary image is performed with the help of masking as shown in figure 10 c). The eroding and dilating process is subjected to the thresholded image to close the areas of missing pixels. During the eroding process, there will be a reduction in the corresponding image by one pixel, while during the dilation process, there will be an increment in the corresponding image by one pixel as shown in figure 10 d) and figure 10 e). The kernel size (3, 3) of type int 8 is created to carry out the eroding and dilating process. The regions in the weld shape are labeled in the masked image as shown in Figure 10 f). To define the nature of pixel connection, i.e., whether the pixels are connected or disconnected, the structure factor of [[1,1,1], [1,1,1], [1,1,1]] is implemented.

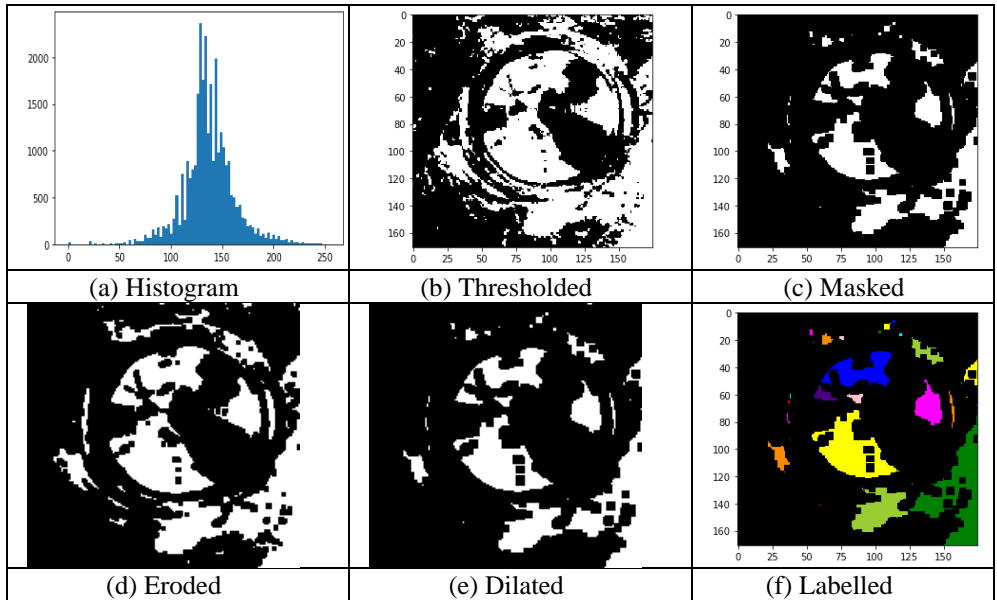

| (a) Histogram | (b) Thresholded | (c) Masked |
| (d) Eroded | (e) Dilated | (f) Labelled |

Figure 10: Image processing results of the sixth weld sample

The geometrical features of the weld sample are done by the extraction of the property from each weld shape region. The obtained geometrical features for each region are shown in table 4.

Table 4. Geometrical Features of the weld sample 6

| Sample Region | Area | Equivalent Diameter | Orientation | Major Axis Length | Minor Axis Length | Perimeter |
| --- | --- | --- | --- | --- | --- | --- |
| 1 | 0.75 | 0.98 | 90.00 | 1.63 | 0.00 | 0.50 |
| 2 | 1.00 | 1.13 | 45.00 | 1.00 | 1.00 | 2.00 |
| 3 | 4.25 | 2.33 | 45.00 | 2.73 | 2.17 | 6.00 |
| 4 | 2.25 | 1.69 | 13.28 | 2.35 | 1.25 | 4.10 |
| 5 | 1.00 | 1.13 | 45.00 | 1.00 | 1.00 | 2.00 |
| 6 | 0.25 | 0.56 | 45.00 | 0.00 | 0.00 | 0.00 |
| 7 | 10.50 | 3.66 | 61.45 | 4.01 | 3.60 | 12.12 |
| 8 | 1.00 | 1.13 | 45.00 | 1.00 | 1.00 | 2.00 |
| 9 | 2.50 | 1.78 | 71.78 | 2.27 | 1.39 | 4.10 |
| 10 | 36.75 | 6.84 | 57.81 | 14.86 | 5.04 | 43.00 |
| 11 | 0.25 | 0.56 | 45.00 | 0.00 | 0.00 | 0.00 |
| 12 | 161.00 | 14.32 | -78.85 | 27.00 | 11.94 | 92.44 |
| 13 | 17.25 | 4.69 | -13.48 | 9.55 | 5.85 | 27.85 |
| 14 | 93.50 | 10.91 | 11.84 | 13.74 | 9.54 | 44.64 |
| 15 | 0.50 | 0.80 | 0.00 | 1.00 | 0.00 | 0.00 |
| 16 | 23.75 | 5.50 | 70.05 | 8.78 | 4.11 | 22.88 |
| 17 | 9.25 | 3.43 | 5.05 | 14.07 | 1.31 | 17.91 |
| 18 | 1.00 | 1.13 | 0.00 | 2.24 | 0.00 | 1.00 |
| 19 | 12.00 | 3.91 | 82.54 | 5.66 | 3.48 | 14.86 |
| 20 | 0.25 | 0.56 | 45.00 | 0.00 | 0.00 | 0.00 |
| 21 | 0.50 | 0.80 | 0.00 | 1.00 | 0.00 | 0.00 |
| 22 | 0.25 | 0.56 | 45.00 | 0.00 | 0.00 | 0.00 |
| 23 | 313.00 | 19.96 | 50.40 | 29.41 | 18.76 | 151.70 |
| 24 | 2.00 | 1.60 | 8.25 | 4.63 | 0.56 | 3.21 |
| 25 | 380.75 | 22.02 | -9.22 | 44.32 | 23.31 | 216.77 |
| 26 | 0.75 | 0.98 | 0.00 | 1.63 | 0.00 | 0.50 |
| 27 | 33.75 | 6.56 | 25.74 | 10.67 | 5.02 | 29.57 |
| 28 | 0.25 | 0.56 | 45.00 | 0.00 | 0.00 | 0.00 |
| 29 | 0.25 | 0.56 | 45.00 | 0.00 | 0.00 | 0.00 |
| 30 | 177.00 | 15.01 | -56.11 | 22.07 | 12.23 | 84.13 |
| 31 | 0.25 | 0.56 | 45.00 | 0.00 | 0.00 | 0.00 |

## 5. CONCLUSIONS

In the present study, Supervised Machine Learning regression-based algorithms such as Support Vector Machine (SVM), Random Forest, and Robust Regression were successfully implemented for the prediction of maximum penetration depth (mm) in the weld samples obtained by the FSSW process. The results showed that the Robust Regression algorithm resulted in the best accuracy for the prediction of maximum penetration depth. The second focus of the recent work was to calculate the geometrical features of the weld samples by using image processing algorithms. The results showed that these algorithms can be considered for the calculation of the area, minor axis length, major axis length, equivalent diameter, and the perimeter of the weld samples.

The future scope of this work can be the further implementation of the Machine Learning based Genetic Algorithms and to further compare the accuracy of the obtained metrics features of Supervised learning-based Machine Learning Regression algorithms.